\documentclass[11pt,letterpaper]{article}
\usepackage{emnlp2017}
\usepackage{times}
\usepackage{url}
\usepackage{latexsym}
\usepackage{graphicx}
\usepackage{epsfig}

\usepackage{amsmath}
\usepackage{amssymb}

\usepackage[linesnumbered,vlined,ruled]{algorithm2e}

\makeatletter
\newcommand\footnoteref[1]{\protected@xdef\@thefnmark{\ref{#1}}\@footnotemark}
\makeatother

\emnlpfinalcopy 

\title{Can string kernels pass the test of time in Native Language Identification?}

\author{Radu Tudor Ionescu{\color{darkblue}$^*$} \and Marius Popescu\thanks{\hspace*{0.8em}The authors have equally contributed to this work.}\\
  \\
  University of Bucharest\\
  Department of Computer Science\\
  14 Academiei, Bucharest, Romania\\
  {\tt raducu.ionescu@gmail.com}\\
  {\tt popescunmarius@gmail.com}  
}

\date{}

\begin{document}
\maketitle
\begin{abstract}
We describe a machine learning approach for the 2017 shared task on Native Language Identification (NLI). The proposed approach combines several kernels using multiple kernel learning. While most of our kernels are based on character $p$-grams (also known as $n$-grams) extracted from essays or speech transcripts, we also use a kernel based on i-vectors, a low-dimensional representation of audio recordings, provided by the shared task organizers. For the learning stage, we choose Kernel Discriminant Analysis (KDA) over Kernel Ridge Regression (KRR), because the former classifier obtains better results than the latter one on the development set. In our previous work, we have used a similar machine learning approach to achieve state-of-the-art NLI results. The goal of this paper is to demonstrate that our shallow and simple approach based on string kernels (with minor improvements) can pass the test of time and reach state-of-the-art performance in the 2017 NLI shared task, despite the recent advances in natural language processing. We participated in all three tracks, in which the competitors were allowed to use only the essays (essay track), only the speech transcripts (speech track), or both (fusion track). Using only the data provided by the organizers for training our models, we have reached a macro $F_1$ score of $86.95\%$ in the closed essay track, a macro $F_1$ score of $87.55\%$ in the closed speech track, and a macro $F_1$ score of $93.19\%$ in the closed fusion track. With these scores, our team (UnibucKernel) ranked in the first group of teams in all three tracks, while attaining the best scores in the speech and the fusion tracks.
\end{abstract}

\section{Introduction}
\label{intro}

Native Language Identification (NLI) is the task of identifying the native language (L1) of a person, based on a sample of text or speech they have produced in a language (L2) other than their mother tongue. This is an interesting sub-task in forensic linguistic applications such as plagiarism detection and authorship identification, where the native language of an author is just one piece of the puzzle \cite{estival-et-al:07}. NLI can also play a key role in second language acquisition (SLA) applications where NLI techniques are used to identify language transfer patterns that help teachers and students focus feedback and learning on particular areas of interest \cite{rozovskaya-roth:2010:EMNLP,jarvis-2012}. 

In 2013, \newcite{tetreault-blanchard-cahill:2013:BEA8} organized the first NLI shared task, providing the participants written essays of non-native English learners. In 2016, the Computational Paralinguistics Challenge~\cite{compare2016} included a shared task on NLI based on the spoken response of non-native English speakers. The 2017 NLI shared task~\cite{nli2017} attempts to combine these approaches by including a written response (essay) and a spoken response (speech transcript and i-vector acoustic features) for each subject. Our team (UnibucKernel) participated in all three tracks proposed by the organizers of the 2017 NLI shared task, in which the competitors were allowed to use only the essays (closed essay track), only the speech transcripts (closed speech track), or both modalities (closed fusion track).

Our approach in each track combines two or more kernels using multiple kernel learning. The first kernel that we considered is the $p$-grams presence bits kernel\footnote{We computed the $p$-grams presence bits kernel using the  code available at http://string-kernels.herokuapp.com.}, which takes into account only the presence of $p$-grams instead of their frequency. The second kernel is the (histogram) intersection string kernel\footnote{We computed the intersection string kernel using the code available at http://string-kernels.herokuapp.com.}, which was first used in a text mining task by \newcite{ionescu-popescu-cahill-EMNLP-2014}. While these kernels are based on character $p$-grams extracted from essays or speech transcrips, we also use an RBF kernel~\cite{taylor-Cristianini-cup-2004} based on i-vectors~\cite{Dehak-IS-2011}, a low-dimensional representation of audio recordings, made available by the 2017 NLI shared task organizers~\cite{nli2017}. We have also considered squared RBF kernel versions of the string kernels and the kernel based on i-vectors. We have taken into consideration two kernel classifiers~\cite{taylor-Cristianini-cup-2004} for the learning task, namely Kernel Ridge Regression (KRR) and Kernel Discriminant Analysis (KDA). 
In a set of preliminary experiments performed on the development set, we found that KDA gives better results than KRR, which is consistent with our previous work~\cite{ionescu-popescu-cahill-EMNLP-2014,ionescu-popescu-cahill-COLI-2016}. Therefore, we decided to submit results using just the KDA classifier. We have also tuned the range of $p$-grams for the string kernels. Using only the data provided by the organizers for training our models, we have reached a weighted $F_1$ score of $86.95\%$ in the essay track, a weighted $F_1$ score of $87.55\%$ in the speech track, and a weighted $F_1$ score of $93.19\%$ in the fusion track. 

The first time we used string kernels for NLI, we placed third in the 2013 NLI shared task~\cite{popescu-ionescu:2013:BEA8}. In 2014, we improved our method and reached state-of-the-art performance~\cite{ionescu-popescu-cahill-EMNLP-2014}. More recently, we have shown that our method is language independent and robust to topic bias~\cite{ionescu-popescu-cahill-COLI-2016}. However, with all the improvements since 2013, our method remained a simple and shallow approach. In spite of its simplicity, the aim of this paper is to demonstrate that our approach can still achieve state-of-the-art NLI results, 4 years after its conception.

The paper is organized as follows. Related work on native language identification and string kernels is presented in Section~\ref{sec_Related_Work}. Section~\ref{sec_Similarity_Measures} presents the kernels that we used in our approach. The learning methods used in the experiments are described in Section~\ref{sec_Learning_Methods}. Details about the NLI experiments are provided in Section~\ref{sec_Experiments}. Finally, we draw conclusions and discuss future work in Section~\ref{sec_Conclusion}.

\section{Related Work}
\label{sec_Related_Work}

\subsection{Native Language Identification}
\label{sec_Related_Work_NLI}

As defined in the introduction, the goal of automatic native language identification (NLI) is to determine the native language of a language learner, based on a piece of writing or speech in a foreign language. 
Most research has focused on identifying the native language of English language learners, though there have been some efforts recently to identify the native language of writing in other languages, such as Chinese~\cite{malmasi-dras:2014:EACL2014-SP} or Arabic~\cite{malmasi-ANLP-2014}.

The first work to study automated NLI was that of~\newcite{Tomokiyo-2001}. In their study, a Na\"{\i}ve Bayes model is trained to distinguish speech transcripts produced by native versus non-native English speakers. A few years later, a second study on NLI appeared~\cite{Jarvis-2004}. In their work, \newcite{Jarvis-2004} tried to determine how well a Discriminant Analysis classifier could predict the L1 language of nearly five hundred English learners from different backgrounds. To make the task more challenging, they included pairs of closely related L1 languages, such as Portuguese and Spanish. The seminal paper by~\newcite{koppel:2005:LNCS} introduced some of the best-performing features for the NLI task: character, word and part-of-speech $n$-grams along with features inspired by the work in the area of second language acquisition such as spelling and grammatical errors. In general, most approaches to NLI have used multi-way classification with SVM or similar models along with a range of linguistic features. The book of~\newcite{jarvis-2012} presents some of the state-of-the-art approaches used up until 2012. Being the first book of its kind, it focuses on the automated detection of L2 language-use patterns that are specific to different L1 backgrounds, with the help of text classification methods. Additionally, the book presents methodological tools to empirically test language transfer hypotheses, with the aim of explaining how the languages that a person knows interact in the mind.

In 2013, \newcite{tetreault-blanchard-cahill:2013:BEA8} organized the first shared task in the field. This allowed researchers to compare approaches for the first time on a specifically designed NLI corpus that was much larger than previously available data sets. In the shared task, $29$ teams submitted results for the test set, and one of the most successful aspects of the competition was that it drew submissions from teams working in a variety of research fields. The submitted systems utilized a wide range of machine learning approaches, combined with several innovative feature contributions. The best performing system in the closed task achieved an overall accuracy of $83.6\%$ on the 11-way classification of the test set, although there was no significant difference between the top teams. Since the 2013 NLI shared task, several systems~\cite{bykh-2014,bykh-meurers:2016:COLING, ionescu-popescu-cahill-EMNLP-2014,ionescu-popescu-cahill-COLI-2016} have reported results above the top scoring system of the 2013 NLI shared task.

Another interesting linguistic interpretation of native language identification data was only recently addressed, specifically the analysis of second language usage patterns caused by native language interference. Usually, language transfer is studied by Second Language Acquisition researchers using manual tools. Language transfer analysis based on automated native language identification methods has been the approach of~\newcite{jarvis-2012}. \newcite{Swanson-EACL-2014} also define a computational methodology that produces a ranked list of syntactic patterns that are correlated with language transfer. Their methodology allows the detection of fairly obvious language transfer effects, without being able to detect underused patterns. The first work to address the automatic extraction of underused and overused features on a per native language basis is that of~\newcite{malmasi-EMNLP-2014}. The work of~\newcite{ionescu-popescu-cahill-COLI-2016} also addressed the automatic extraction of underused and overused features captured by character $p$-grams.

\subsection{String Kernels}

In recent years, methods of handling text at the character level have demonstrated impressive performance levels in various text analysis tasks~\cite{LodhiSSCW02,Sanderson:06,Kate-ACL-2006,encoplot,popescu:2011:RANLP,Escalante-ACL-2011,PopescuG12,popescu-ionescu:2013:BEA8,ionescu-popescu-cahill-EMNLP-2014,ionescu-popescu-cahill-COLI-2016, franco-EACL-2017, Radu-Andrei-ADI-2017}. String kernels are a common form of using information at the character level. They are a particular case of the more general convolution kernels~\cite{haussler-1999}. \newcite{LodhiSSCW02} used string kernels for document categorization with very good results. String kernels were also successfully used in authorship identification~\cite{Sanderson:06,PopescuG12}. For example, the system described by~\newcite{PopescuG12} ranked first in most problems and overall in the PAN 2012 Traditional Authorship Attribution tasks. More recently, various blended string kernels reached state-of-the-art accuracy rates for native language identification~\cite{ionescu-popescu-cahill-EMNLP-2014,ionescu-popescu-cahill-COLI-2016} and Arabic dialect identification~\cite{Radu-Marius-ADI-2016, Radu-Andrei-ADI-2017}. String kernels have also been used for sentiment analysis in various languages~\cite{marius-KES-2017} and in cross-domain settings~\cite{franco-EACL-2017}.  

\section{Kernels for Native Language Identification}
\label{sec_Similarity_Measures}


\subsection{String Kernels}
\label{sec_String_Kernels}

The kernel function captures the intuitive notion of similarity between objects in a specific domain and can be any function defined on the respective domain that is symmetric and positive definite. For strings, many such kernel functions exist with various applications in computational biology and computational linguistics~\cite{taylor-Cristianini-cup-2004}. String kernels embed the texts in a very large feature space, given by all the substrings of length $p$, and leave it to the learning algorithm to select important features for the specific task, by highly weighting these features.

The first kernel that we use in the NLI experiments is the character $p$-grams presence bits kernel. The feature map defined by this kernel associates to each string a vector of dimension $|\Sigma|^p$ containing the presence bits of all its substrings of length $p$ ($p$-grams). Formally, for two strings over an alphabet $\Sigma$, $s,t \in \Sigma^*$, the character $p$-grams presence bits kernel is defined as:
\begin{equation*}
\begin{split}
k^{0/1}_p(s,t)=\sum\limits_{v \in \Sigma^p}\mbox{in}_v(s) \cdot \mbox{in}_v(t),
\end{split}
\end{equation*}
where $\mbox{in}_v(s)$ is $1$ if string $v$ occurs as a substring in $s$, and $0$ otherwise.

The second kernel that we employ is the intersection string kernel introduced in~\cite{ionescu-popescu-cahill-EMNLP-2014}. The intersection string kernel is defined as follows:
\begin{equation*}
\begin{split}
k^{\cap}_p(s,t)=\sum\limits_{v \in \Sigma^p} \min \lbrace \mbox{num}_v(s), \mbox{num}_v(t) \rbrace ,
\end{split}
\end{equation*}
where $\mbox{num}_v(s)$ is the number of occurrences of string $v$ as a substring in $s$. Further details about the string kernels for NLI are given in~\cite{ionescu-popescu-cahill-COLI-2016}. The efficient algorithm for computing the string kernels is presented in~\cite{marius-KES-2017}.

Data normalization helps to improve machine learning performance for various applications. Since the value range of raw data can have large variation, classifier objective functions will not work properly without normalization. 
After normalization, each feature has an approximately equal contribution to the similarity between two samples.
%
To ensure a fair comparison of strings of different lengths, normalized versions of the $p$-grams presence bits kernel and the intersection kernel are being used:
\begin{equation*}
\begin{split}
\hat{k}^{0/1}_p(s,t) & =\frac{k^{0/1}_p(s,t)}{\sqrt{k^{0/1}_p(s,s) \cdot k^{0/1}_p(t,t)}},\\
\hat{k}^{\cap}_p(s,t) & =\frac{k^{\cap}_p(s,t)}{\sqrt{k^{\cap}_p(s,s) \cdot k^{\cap}_p(t,t)}}.
\end{split}
\end{equation*}
Taking into account $p$-grams of different lengths and summing up the corresponding kernels,
new kernels, termed \emph{blended spectrum kernels}, can be obtained. We have used various blended spectrum kernels in the experiments in order to find the best combination. Inspired by the success of~\newcite{Radu-Andrei-ADI-2017} in using a squared RBF kernel based on i-vectors for Arabic dialect identification, we have also tried out squared RBF versions of the above kernels. We first compute the standard RBF kernels as follows:
\begin{equation*}
\begin{split}
\bar{k}^{0/1}_p(s,t) & = exp \left( - \frac{\displaystyle 1 - \hat{k}^{0/1}_p(s,t)} {\displaystyle 2 \sigma^2} \right),\\
\bar{k}^{\cap}_p(s,t) & = exp \left( - \frac{\displaystyle 1 - \hat{k}^{\cap}_p(s,t)} {\displaystyle 2 \sigma^2} \right).
\end{split}
\end{equation*}

We then interpret the RBF kernel matrix as a feature matrix, and apply the dot product to obtain a linear kernel for this new representation:
\begin{equation*}
\bar{K} = K \cdot K'.
\end{equation*}

The resulted squared RBF kernels are denoted by $(\bar{k}^{0/1}_p)^2$ and $(\bar{k}^{\cap}_p)^2$, respectively.


\subsection{Kernel based on Acoustic Features}

For the speech and the fusion tracks, we also build a kernel from the i-vectors provided by the organizers~\cite{nli2017}. The i-vector approach~\cite{Dehak-IS-2011} is a powerful speech modeling technique that comprises all the updates happening during the adaptation of a Gaussian mixture model (GMM) mean components to a given utterance. The provided i-vectors have $800$ dimensions. In order to build a kernel from the i-vectors, we first normalize the i-vectors using the $L_2$-norm, then we compute the euclidean distance between each pair of i-vectors. We next employ the RBF kernel~\cite{taylor-Cristianini-cup-2004} to transform the distance into a similarity measure:
\begingroup
\large
\begin{equation*}
\hat{k}^{\textit{i-vec}}(x, y) = exp \left(- \frac{\displaystyle \sqrt{\sum^{m}_{j=1}(x_j - y_j)^2}} {\displaystyle 2 \sigma^2} \right),
\end{equation*}
\endgroup
where $x$ and $y$ are two i-vectors and $m$ represents the size of the two i-vectors, $800$ in our case. For optimal results, we have tuned the parameter $\sigma$ in a set of preliminary experiments. We also interpret the resulted similarity matrix as a feature matrix, and we compute the product between the matrix and its transpose to obtain the squared RBF kernel based on i-vectors, denoted by $(\bar{k}^{\textit{i-vec}})^2$.

\section{Learning Methods}
\label{sec_Learning_Methods}

Kernel-based learning algorithms work by embedding the data into a Hilbert feature space and by searching for linear relations in that space. The embedding is performed implicitly, by specifying the inner product between each pair of points rather than by giving their coordinates explicitly. More precisely, a kernel matrix that contains the pairwise similarities between every pair of training samples is used in the learning stage to assign a vector of weights to the training samples.


Various kernel methods differ in the way they learn to separate the samples. In the case of binary classification problems, kernel-based learning algorithms look for a discriminant function, a function that assigns $+1$ to examples belonging to one class and $-1$ to examples belonging to the other class. In the NLI experiments, we employed the Kernel Ridge Regression (KRR) binary classifier. Kernel Ridge Regression selects the vector of weights that simultaneously has small empirical error and small norm in the Reproducing Kernel Hilbert Space generated by the kernel function. 
KRR is a binary classifier, but native language identification is usually a multi-class classification problem. There are many approaches for combining binary classifiers to solve multi-class problems. Typically, the multi-class problem is broken down into multiple binary classification problems using common decomposition schemes such as: one-versus-all and one-versus-one. We considered the one-versus-all scheme for our NLI task. There are also kernel methods that take the multi-class nature of the problem directly into account, for instance Kernel Discriminant Analysis. The KDA classifier is sometimes able to improve accuracy by avoiding the masking problem~\cite{ESL-hastie-tibshirani-2003}. 
More details about the kernel classifiers employed for NLI are discussed in~\cite{ionescu-popescu-cahill-COLI-2016}.

\section{Experiments}
\label{sec_Experiments}

\subsection{Data Set}

The corpus provided for the 2017 NLI shared task contains 13,200 multi-modal samples produces by speakers of the following 11 languages: Arabic, Chinese, French, German, Hindi, Italian, Japanese, Korean, Spanish, Telugu and Turkish. The samples are split into 11,000 for training, 1100 for development and 1100 for testing. The distribution of samples per prompt (topic) per native language is balanced. Each sample is composed of an essay and an audio recording of a non-native English learner. For privacy reasons, the shared task organizers were not able to provide the original audio recordings. Instead, they provided a speech transcript and an i-vector representation derived from the audio file.

\subsection{Parameter and System Choices}

In our approach, we treat essays or speech transcripts as strings. Because the approach works at the character level, there is no need to split the texts into words, or to do any NLP-specific processing before computing the string kernels. Hence, we apply string kernels on the raw text samples, disregarding the tokenized version of the samples. The only editing done to the texts was the replacing of sequences of consecutive space characters (space, tab, and so on) with a single space character. This normalization was needed in order to prevent the artificial increase or decrease of the similarity between texts, as a result of different spacing.

\begin{table}
\centering
\begin{tabular}{|l|c|c|}
\hline
\bf Kernel 																										& \multicolumn{2}{c|}{\bf Accuracy}\\
\cline{2-3}
\bf 																													& \bf KRR			& \bf KDA\\
\hline
$\hat{k}^{0/1}_{5-9}$ 																					& $82.18\%$ 	& $84.55\%$ \\
$\hat{k}^{\cap}_{5-9}$ 																				& $81.91\%$	& $84.18\%$ \\
\hline
\end{tabular}
\caption{Accuracy rates of KRR versus KDA on the essay development set.}
\label{tab_KRR_vs_KDA}
\vspace*{-0.8em}
\end{table}

We used the development set for tuning the parameters of our approach. Although we have some intuition from our previous work~\cite{ionescu-popescu-cahill-COLI-2016} about the optimal range of $p$-grams that can be used for NLI from essays, we decided to carry out preliminary experiments in order to confirm our intuition. We also carried out preliminary experiments to determine the optimal range of $p$-grams to be used for speech transcripts, a different kind of representation that captures other features of the non-native English speakers. We fixed the learning method to KDA based on the presence bits kernel and we evaluated all the $p$-grams in the range $3$-$9$. For essays, we found that $p$-grams in the range $5$-$9$ work best, which confirms our previous results on raw text documents reported in~\cite{ionescu-popescu-cahill-COLI-2016}. For speech transcripts, we found that longer $p$-grams are not helpful. Thus, the optimal range of $p$-grams is $5$-$7$. In order to decide which classifier gives higher accuracy rates, we carried out some preliminary experiments using only the essays. The KRR and the KDA classifiers are compared in Table~\ref{tab_KRR_vs_KDA}. We observe that KDA yields better results for both the blended $p$-grams presence bits kernel ($\hat{k}^{0/1}_{5-9}$) and the blended $p$-grams intersection kernel ($\hat{k}^{\cap}_{5-9}$). Therefore, we employ KDA for the subsequent experiments. An interesting remark is that we also obtained better performance with KDA instead of KRR for the English L2, in our previous work~\cite{ionescu-popescu-cahill-COLI-2016}.

\begin{table*}[t!]
\centering
\begin{tabular}{|l|c|c|}
\hline

\bf Kernel																											& \bf Accuracy			& \bf Track\\
\hline
$\hat{k}^{0/1}_{5-9}$ 																					& $84.55\%$ 	& Essay \\
$\hat{k}^{\cap}_{5-9}$ 																				& $84.18\%$	& Essay \\
$\hat{k}^{0/1}_{5-9} + \hat{k}^{\cap}_{5-9}$ 											& $\mathbf{85.18\%}$ 	& Essay \\
$(\bar{k}^{0/1}_{5-9})^2$ 																			& $85.45\%$ 	& Essay \\
$(\bar{k}^{\cap}_{5-9})^2$ 																			& $85.09\%$	& Essay \\
$(\bar{k}^{0/1}_{5-9})^2 + (\bar{k}^{\cap}_{5-9})^2$ 							& $\mathbf{85.55\%}$ 	& Essay \\
\hline
$\hat{k}^{0/1}_{5-7}$ 																					& $58.73\%$ 	& Speech \\
$\hat{k}^{\cap}_{5-7}$ 																				& $58.55\%$	& Speech \\
$\hat{k}^{\textit{i-vec}}$																				& $81.64\%$ 	& Speech \\
$\hat{k}^{0/1}_{5-7} + \hat{k}^{\cap}_{5-7}$ 											& $58.73\%$ 	& Speech \\
$\hat{k}^{0/1}_{5-7} + \hat{k}^{\textit{i-vec}}$ 										& $\mathbf{85.27\%}$ 	& Speech \\
$\hat{k}^{\cap}_{5-7} + \hat{k}^{\textit{i-vec}}$ 										& $85.18\%$ 	& Speech \\
$\hat{k}^{0/1}_{5-7} + \hat{k}^{\cap}_{5-7} + \hat{k}^{\textit{i-vec}}$ 	& $84.91\%$ 	& Speech \\

$(\bar{k}^{0/1}_{5-7})^2$ 																			& $59.00\%$ 	& Speech \\
$(\bar{k}^{\cap}_{5-7})^2$ 																			& $59.82\%$	& Speech \\
$(\bar{k}^{\textit{i-vec}})^2$																		& $81.55\%$ 	& Speech \\
$(\bar{k}^{0/1}_{5-7})^2 + (\bar{k}^{\cap}_{5-7})^2$ 							& $59.91\%$ 	& Speech \\
$(\bar{k}^{0/1}_{5-7})^2 + (\bar{k}^{\textit{i-vec}})^2$ 							& $85.36\%$ 	& Speech \\
$(\bar{k}^{\cap}_{5-7})^2 + (\bar{k}^{\textit{i-vec}})^2$ 						& $85.27\%$ 	& Speech \\
$(\bar{k}^{0/1}_{5-7})^2 + (\bar{k}^{\cap}_{5-7})^2 + (\bar{k}^{\textit{i-vec}})^2$ & $\mathbf{85.45\%}$ 	& Speech \\
\hline
$\hat{k}^{0/1}_{5-9} + \hat{k}^{\cap}_{5-9} + \hat{k}^{0/1}_{5-7} + \hat{k}^{\textit{i-vec}}$
																														& $91.64\%$ 	& Fusion \\
																														
$\hat{k}^{0/1}_{5-9} + \hat{k}^{0/1}_{5-7} + \hat{k}^{\textit{i-vec}}$	& $\mathbf{92.09\%}$ 	& Fusion \\
																														
$(\bar{k}^{0/1}_{5-9})^2 + (\bar{k}^{\cap}_{5-9})^2 + (\bar{k}^{0/1}_{5-7})^2 + (\bar{k}^{\cap}_{5-7})^2 + (\bar{k}^{\textit{i-vec}})^2$ 										
																														& $\mathbf{91.72\%}$ 	& Fusion \\
\hline
\end{tabular}
\caption{Accuracy rates on the NLI development set obtained by KDA based on various kernels for the essay, the speech and the fusion tracks. The submitted systems are highlighted in bold.}
\label{tab_dev_results}
\vspace*{-0.1em}
\end{table*}

After fixing the classifier and the range of $p$-grams for each modality, we conducted further experiments to establish what type of kernel works better, namely the blended $p$-grams presence bits kernel, the blended $p$-grams intersection kernel, or the kernel based on i-vectors. We also included squared RBF versions of these kernels. Since these different kernel representations are obtained either from essays, speech transcripts or from low-level audio features, a good approach for improving the performance is combining the kernels. When multiple kernels are combined, the features are actually embedded in a higher-dimensional space. As a consequence, the search space of linear patterns grows, which helps the classifier in selecting a better discriminant function. The most natural way of combining two or more kernels is to sum them up. Summing up kernels or kernel matrices is equivalent to feature vector concatenation. The kernels were evaluated alone and in various combinations, by employing KDA for the learning task. All the results obtained on the development set are given in Table~\ref{tab_dev_results}.

\begin{table*}[t!]
\centering
\begin{tabular}{|l|c|c|c|c|}
\hline

\bf Kernel																											
					& \bf Accuracy			& \bf $\mathbf{F_1}$ (macro) 	& \bf Track & \bf Rank \\
\hline
$\hat{k}^{0/1}_{5-9} + \hat{k}^{\cap}_{5-9}$ 											
					& $86.91\%$ 				& $\mathbf{86.95\%}$				& Essay & 1st\\
$(\bar{k}^{0/1}_{5-9})^2 + (\bar{k}^{\cap}_{5-9})^2$ 							
					& $86.91\%$ 				& $\mathbf{86.95\%}$ 				& Essay & 1st\\
\hline
$\hat{k}^{0/1}_{5-7} + \hat{k}^{\textit{i-vec}}$ 										
					& $87.55\%$ 				& $\mathbf{87.55\%}$ 				& Speech & 1st\\
$(\bar{k}^{0/1}_{5-7})^2 + (\bar{k}^{\cap}_{5-7})^2 + (\bar{k}^{\textit{i-vec}})^2$ 
					& $87.45\%$ 				& $87.45\%$								& Speech & 1st\\
\hline
$\hat{k}^{0/1}_{5-9} + \hat{k}^{0/1}_{5-7} + \hat{k}^{\textit{i-vec}}$
					& $93.18\%$ 				& $\mathbf{93.19\%}$				& Fusion & 1st\\
																														
$(\bar{k}^{0/1}_{5-9})^2 + (\bar{k}^{\cap}_{5-9})^2 + (\bar{k}^{0/1}_{5-7})^2 + (\bar{k}^{\cap}_{5-7})^2 + (\bar{k}^{\textit{i-vec}})^2$ 										
					& $93.00\%$ 				& $93.01\%$								& Fusion & 1st\\
\hline
\end{tabular}
\caption{Accuracy rates on the NLI test set obtained by KDA based on various kernels for the essay, the speech and the fusion tracks. The best marco $F_1$ score in each track is highlighted in bold. The final rank of each kernel combination in the 2017 NLI shared task is presented on the last column.}
\label{tab_test_results}
\end{table*}

The empirical results presented in Table~\ref{tab_dev_results} reveal several interesting patterns of the proposed methods. On the essay development set, the presence bits kernel gives slightly better results than the intersection kernel. The combined kernels yield better performance than each of the individual components, which is remarkably consistent with our previous works~\cite{ionescu-popescu-cahill-EMNLP-2014, ionescu-popescu-cahill-COLI-2016}. For each kernel, we obtain an improvement of up to $1\%$ by using the squared RBF version. The best performance on the essay development set ($85.55\%$) is obtained by sum of the squared RBF presence bits kernel and the squared RBF intersection kernel. On the speech track, the results are fairly similar among the string kernels, but the kernel based on i-vectors definitely stands out. Indeed, the best individual kernel is the kernel based on i-vectors with an accuracy of $81.64\%$. By contrast, the best individual string kernel is the squared RBF intersection kernel, which yields an accuracy of $59.82\%$. Thus, it seems that the character $p$-grams extracted from speech transcripts do not provide enough information to accurately distinguish the native languages. On the other hand, the i-vector representation extracted from audio recordings is much more suitable for the NLI task. Interestingly, we obtain consistently better results when we combine the kernels based on i-vectors with one or both of the string kernels. The best performance on the speech development set ($85.45\%$) is obtained by sum of the squared RBF presence bits kernel, the squared RBF intersection kernel and the squared RBF kernel based on i-vectors. The top accuracy levels on the essay and speech development sets are remarkably close. Nevertheless, when we fuse the features captured by the kernels constructed for the two modalities, we obtain considerably better results. This suggests that essays and speech provide complementary information, boosting the accuracy of the KDA classifier by more than $6\%$ on the fusion development set. It is important to note that we tried to fuse the kernel combinations that provided the best performance on the essay and the speech development sets, while keeping the original and the squared RBF versions separated. We also tried out a combination that does not include the intersection string kernel, an idea that seems to improve the performance. Actually, the best performance on the fusion development set ($92.09\%$) is obtained by sum of the presence bits kernel ($\hat{k}^{0/1}_{5-9}$) computed from essays, the presence bits kernel ($\hat{k}^{0/1}_{5-7}$) computed from speech transcripts, and the kernel based on i-vectors ($\hat{k}^{\textit{i-vec}}$). In each track, we submitted the top two kernel combinations for the final test evaluation.

\subsection{Results}

\begin{figure*}[t!]
\centering
\includegraphics[width=0.7\textwidth]{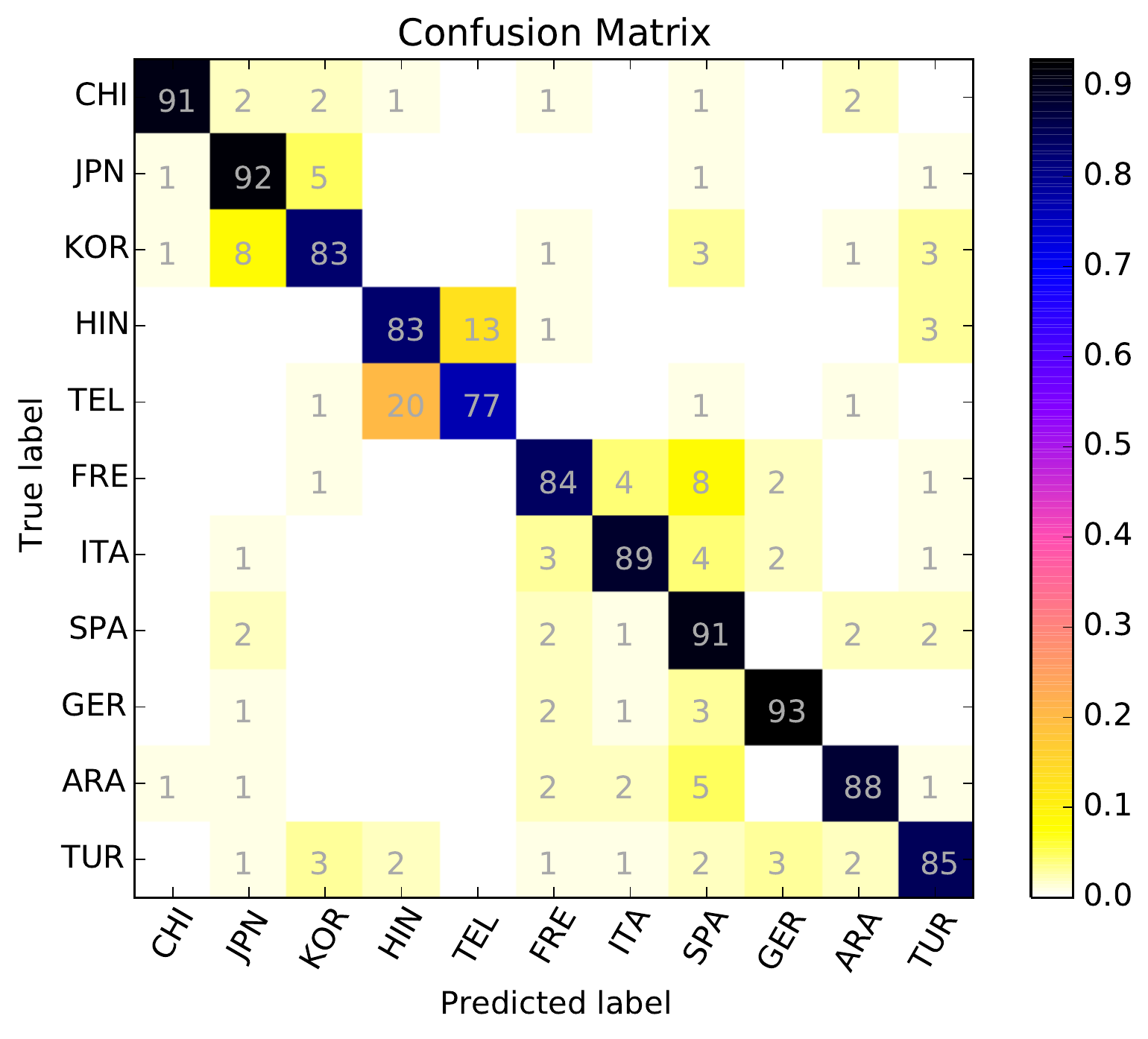}
\vspace*{-1.3em}
\caption{Confusion matrix of the system based on squared RBF kernels on the essay track.}
\label{fig_essay}
\vspace*{-0.2em}
\end{figure*}

\begin{figure*}[t!]
\centering
\includegraphics[width=0.7\textwidth]{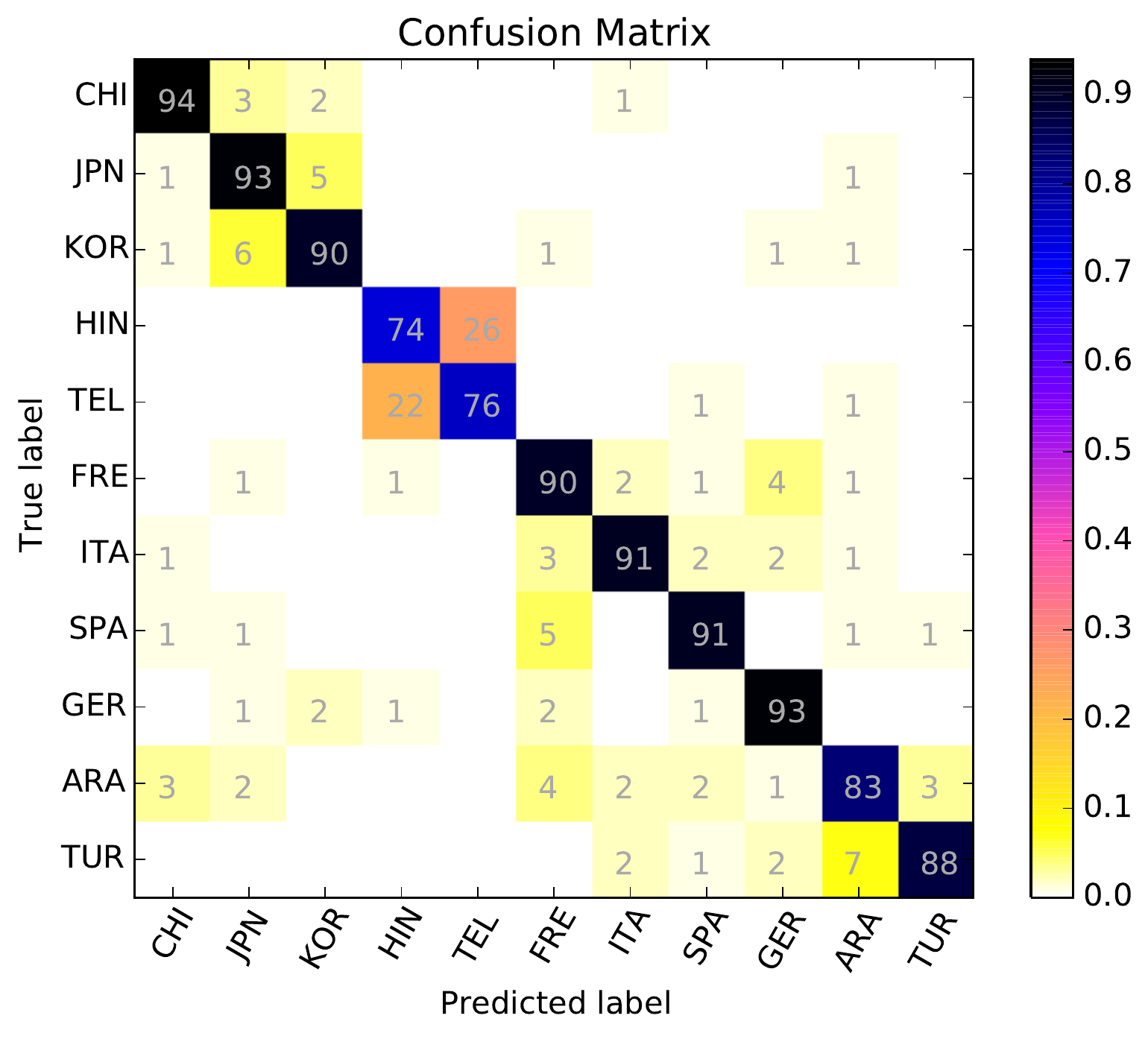}
\vspace*{-1.3em}
\caption{Confusion matrix of the best system on the speech track.}
\label{fig_speech}
\end{figure*}

The results on the test set are presented in Table~\ref{tab_test_results}. Although we tuned our approach to optimize the accuracy rate, the official evaluation metric for the NLI task is the macro $F_1$ score. Therefore, we have reported both the accuracy rate and the macro $F_1$ score in Table~\ref{tab_test_results}. Both kernel combinations submitted to the essay track obtain equally good results ($86.95\%$). For the speech and the fusion tracks, the squared RBF kernels reach slightly lower performance than the original kernels. The best submission to the speech track is the KDA classifier based on the sum of the presence bits kernel ($\hat{k}^{0/1}_{5-7}$) and the kernel based on i-vectors ($\hat{k}^{\textit{i-vec}}$), a combination that reaches a macro $F_1$ score of $87.55\%$. These two kernels are also included in the sum of kernels that gives our top performance in the fusion track ($93.19\%$). Along with the two kernels, the best submission to the fusion track also includes the presence bits kernel ($\hat{k}^{0/1}_{5-9}$) computed from essays. An interesting remark is that the results on the test set (Table~\ref{tab_test_results}) are generally more than $1\%$ better than the results on the development set (Table~\ref{tab_dev_results}), perhaps because we have included the development samples in the training set for the final test evaluation. 

\begin{figure*}[t!]
\centering
\includegraphics[width=0.7\textwidth]{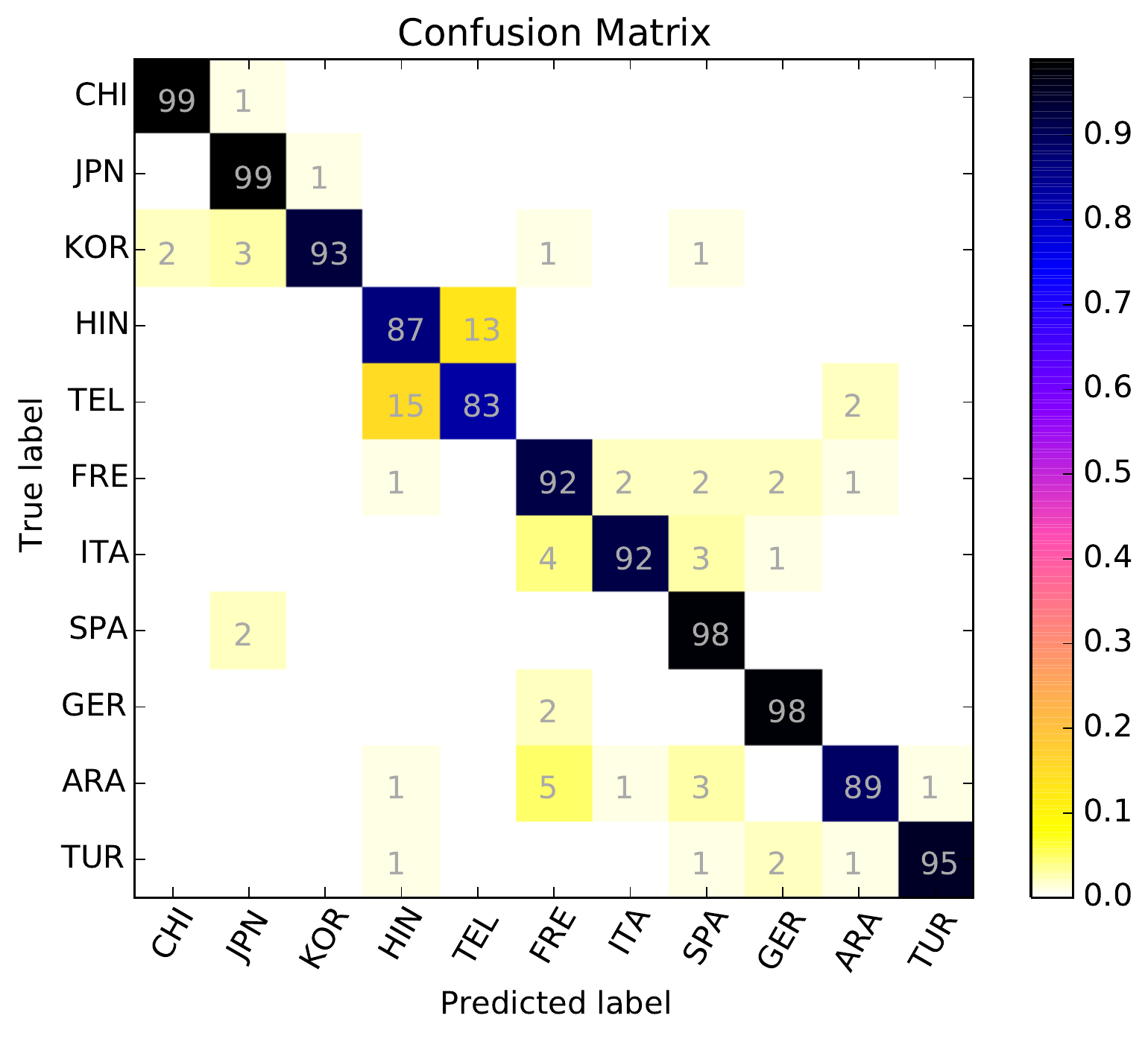}
\vspace*{-1.3em}
\caption{Confusion matrix of the best system on the fusion track.}
\label{fig_fusion}
\vspace*{-0.5em}
\end{figure*}

The organizers have grouped the teams based on statistically significant differences between each team's best submission, calculated using McNemar's test with an alpha value of $0.05$. The macro $F_1$ score of $86.95\%$ places us in the first group of methods in the essay track, although we reach only the sixth best performance within the group. Remarkably, we also rank in the first group of methods in the speech and the fusion tracks, while also reaching the best performance in each of these two tracks. It is important to note that UnibucKernel is the only team ranked in first group of teams in each and every track of the 2017 NLI shared task, indicating that our shallow and simple approach is still state-of-the-art in the field.

To better visualize our results, we have included the confusion matrices for our best runs in each track. The confusion matrix presented in Figure~\ref{fig_essay} shows that our approach for the essay track has a higher misclassification rate for Telugu, Hindi and Korean, while the confusion matrix shown in Figure~\ref{fig_speech} indicates that our approach for the speech track has a higher misclassification rate for Hindi, Telugu and Arabic. Finally, the confusion matrix illustrated in Figure~\ref{fig_fusion}, shows that we are able to obtain the highest correct classification rate for each and every L1 language (with respect to the other two confusion matrices) by fusing the essay and speech information. While there are no more than two misclassified samples for Chinese, Japanese, Spanish and German, our fusion-based approach still has some trouble in distinguishing Hindi and Telugu. Another interesting remark is that $5$ native Arabic speakers are wrongly classified as French, perhaps because these Arabic speakers are from Maghreb, a region in which French arrived as a colonial language. As many people in this region speak French as a second language, it seems that our system gets confused by the mixed Arabic (L1) and French (L2) language transfer patterns that are observable in English (L3).

\section{Conclusion and Future Work}
\label{sec_Conclusion}

In this paper, we have described our approach based on learning with multiple kernels for the 2017 NLI shared task~\cite{nli2017}. Our approach attained generally good results, consistent with those reported in our previous works~\cite{ionescu-popescu-cahill-EMNLP-2014,ionescu-popescu-cahill-COLI-2016}. Indeed, our team (UnibucKernel) ranked in the first group of teams in all three tracks, while reaching the best marco $F_1$ scores in the speech ($87.55\%$) and the fusion ($93.19\%$) tracks. As we are the only team that ranked in first group of teams in each and every track of the 2017 NLI shared task, we consider that our approach has passed the test of time in native language identification.

Although we refrained from including other types of features in order to keep our approach shallow and simple, and to prove that we can achieve state-of-the-art results using character $p$-grams alone, we will consider combining string kernels with other features in future work.

\section*{Acknowledgments}
This research is supported by University of Bucharest, Faculty of Mathematics and Computer Science, through the 2017 Mobility Fund.

\bibliography{nli2017}
\bibliographystyle{emnlp_natbib}

\end{document}